\documentclass[runningheads]{llncs}
\usepackage{amssymb,amsmath}
\usepackage{graphicx}
\usepackage{xcolor}
\usepackage{verbatim}
\usepackage{floatrow}
\usepackage{subcaption}
\usepackage{booktabs}
\usepackage{ntheorem}
\newcommand{\Mone}{\texttt{Mi}}
\newcommand{\Mtwo}{\texttt{Mb}}
\newcommand{\DB}{\texttt{DistilBERT}}
\newcommand{\B}{\texttt{BERT}}

\newtheorem{hyp}{Hypothesis} 
\begin{document}
\title{An investigation of structures responsible for gender bias in BERT and DistilBERT }
\titlerunning{Investigation gender bias in BERT and DistilBERT}
%
\author{T. Leteno\inst{1} \and
A. Gourru\inst{1}\and
C. Laclau\inst{2}\and C. Gravier\inst{1}}
\authorrunning{T. Leteno et al.}
%
\institute{Université Jean Monnet Saint-Étienne, CNRS, Institut d
Optique Graduate School, Laboratoire Hubert Curien UMR 5516, F-42023,
SAINT-ÉTIENNE, FRANCE \email{firstname.lastname@univ-st-etienne.fr} \and Télécom Paris, Institut Polytechnique de Paris\\
\email{charlotte.laclau@telecom-paris.fr}}
\maketitle              

\begin{abstract}
    In recent years, large Transformer-based Pre-trained Language Models (PLM) have changed the Natural Language Processing (NLP) landscape, by pushing the performance boundaries of the state-of-the-art on a wide variety of tasks. However, this performance gain goes along with an increase in complexity, and as a result, the size of such models (up to billions of parameters) represents a constraint for their deployment on embedded devices or short-inference time tasks. To cope with this situation, compressed models emerged (e.g. DistilBERT), democratizing their usage in a growing number of applications that impact our daily lives. A crucial issue is the fairness of the predictions made by both PLMs and their distilled counterparts. In this paper, we propose an empirical exploration of this problem by formalizing two questions: (1) Can we identify the neural mechanism(s) responsible for gender bias in BERT (and by extension DistilBERT)? (2) Does distillation tend to accentuate or mitigate gender bias (e.g. is DistilBERT more prone to gender bias than its \textit{uncompressed version,} BERT)? Our findings are the following: (I) one cannot identify a specific layer that produces bias; (II) every attention head uniformly encodes bias; except in the context of underrepresented classes with a high imbalance of the sensitive attribute; (III) this subset of heads is different as we re-fine tune the network; (IV) bias is more homogeneously produced by the heads in the distilled model.
    
\end{abstract}
\section{Introduction}

The introduction of large Pre-trained Language Models (PLM) has marked an important paradigm shift in Natural Language Processing (NLP). It leads to unprecedented progress in tasks such as machine translation, document classification \cite{devlin2018bert}, and multitasks text generation \cite{radford2019language}. The strength of these approaches lies in their ability to produce contextual representations. 
They have been initially based on Recurrent Neural Networks (RNN) \cite{dai2015} and they have gradually integrated the Transformers model \cite{vaswani2017attention} as is the case for GPT3 \cite{radford2019language} or BERT \cite{devlin2018bert}, for example. Compared to RNNs, Transformers can be parallelized, which opens the way, on one hand, to the use of ever-increasing training corpus (for example, GPT3 is trained on 45TB of data - almost the entire public web), and on the other hand, to the design of increasingly complex architectures (e.g., BERT large comprises 345 million parameters, BERT base 110 million).
In a nutshell, Transformers~\cite{vaswani2017attention} are founded on three key innovations: positional encoding, scaled dot product attention, and multi-head attention (we will come back to these elements in more detail in Section~\ref{sec:background}). As a result of a combination of all these elements, Transformers can learn an internal understanding of language automatically from the data. Despite their good performance on many different tasks, the use of these models in so-called sensitive applications or areas raised concerns over the past couple of years. Indeed, when decisions have an impact on individuals, for example in the medical and legal domains \cite{demner2009can} or human resources \cite{jatoba2019evolution}, it becomes crucial to study the fairness of these models. 

The core definition of fairness is still a hotly debated topic in the scientific community. In our work, we adopt the following commonly accepted definition \cite{mehrabi2021survey}: \textit{fairness refers to the absence of any prejudice or favoritism towards an individual or a group based on their intrinsic or acquired traits}. In machine learning, we assume that \textit{unfairness} is the result of biased predictions (prejudice or favoritism), which are defined as elements that conduct a model to treat groups of individuals conditionally on some particular \textit{protected} attributes, such as gender, race, or sexual orientation. 

As an example, in human resources, the NLP-based recruitment task consists in analyzing and then selecting the relevant candidates. A lack of diversity inherent to the data, for instance, a corpus containing a large majority of male profiles (i.e. sample bias), will cause the model to maintain and accentuate a gender bias~\cite{swinger2019biases}.
When handling simple linear models trained on reasonable size corpora, creating safeguards to avoid this type of bias is conceivable. 
With PLM, the characteristics that allow them to perform so well are numerous: the size of their training corpus, the number of parameters, and their ability to infer a fine-grained semantic from the data. However, they are also what make it difficult to prevent them from encoding societal biases~\cite{bender2021dangers}.

\paragraph{Related Works}

Several recent studies highlight fairness issues raised by models based on the Transformer architecture. %
These issues are observed in different levels of the NLP pipeline: text encoding~\cite{basta2019evaluating,kurita2019measuring}, during the fine-tuning process~\cite{delobell2021measuringfairness}, or simply as the potential harm caused on downstream tasks~\cite{kurita2019measuring}, with dedicated studies on language generation \cite{sheng2019woman}, document classification \cite{bhardwaj2020investigating}, toxicity detection, and sentiment analysis \cite{hutchinson2020social}. Besides measuring the fairness issue, \emph{locating} the neural mechanism responsible for these issues is largely understudied and unsolved -- locating such mechanisms would unlock the possibility for counter-measures in neural architectures. 
At the same time, a segment of research focused on compressing these large pre-trained models to attain similar performances with fewer parameters, so that running these models is more sustainable and more cost-effective. Several model compression techniques have been proposed, as discussed in~\cite{Gupta22tkdd} and namely the following compression families : pruning, quantization and distillation. The primer \cite{lecun1989optimal} increases the speed and generalization capacities by removing the less important model's weights with regard to the task, while quantization approximates the model’s weights to reduce its complexity (e.g. reducing the numerical precision of the weights \cite{kim2021ibert}). Finally, distillation \cite{hinton2015distilling} consists in training a smaller model (called \emph{student model}) to mimic the predictions of the large PLM to distill (\emph{teacher model}). 
In the present work, we focus on this latter, approach.
One of the earliest model, DistilBERT \cite{sanh2020distilbert} is able to reduce the number of parameters of BERT by 40\%  while maintaining 96\% of accuracy in document classification. 
Looking at the impact of model distillation through fairness lenses has started to be investigated, mainly in the context of computer vision~\cite{hooker2020characterising,hooker2021compressed,lukasik2021teacher}. To summarize, their findings: i) compressed models impact underrepresented visual features directly related to bias, and ii) distilled models tend to accentuate discrimination already made by the teacher model. In NLP, fewer works have been conducted, and the conclusions are sometimes contradictory. 
While some works have shown that distilled versions of PLMs can exacerbate bias  \cite{radford2018improving,delobelle2022fairdistillation}, other articles seem to reach an opposite conclusion \cite{xu2022can}; in this latter, authors state that model distillation acts as a regularization technique allowing bias reduction.

\paragraph{Contribution} Based on existing results, we start from the postulate that PLMs, and more specifically BERT, encode undesirable bias. With a focus on the task of document classification on the Bias in Bio dataset, our objective is to identify the inner structure of the neural network architecture that produce bias, both for BERT and its distilled version DistilBERT. To this end, we design and conduct a series of experiments to verify the relation between models' fairness and their intermediate representation or the attention they carry to the embedding in different data balance setting.

\paragraph{Organisation}
Section~\ref{sec:background} provides background knowledge about BERT and DistilBERT. Section~\ref{section:protocol} presents the empirical protocols that we design. Section \ref{section:experiments} details the technical setting and shows the obtained results of our experiments. Finally, we conclude in Section \ref{section:conclusion} and provide several perspectives unlocked by our experiments.

\section{Preliminaries and Background}\label{sec:background}

We study two PLM, BERT~\cite{devlin2018bert} and its distilled version, DistilBERT~\cite{sanh2020distilbert}.
BERT is a general-purpose language model trained on masked language modeling task 
\footnote{The model is also trained on a next sentence prediction task, but that is irrelevant in our work and therefore not presented here.}.
A small fraction of the words of each training document is masked, and the model is trained to reconstruct those masked words on a large amount of textual data. More precisely, the encoder part of a Transformer architecture takes a sentence, or short document, as input, and maps each token (word or subwords) to an initial representation space in $\mathbb{R}^{768}$.
 
\begin{figure}[t]
        \centering
        \includegraphics[width=0.8\textwidth]{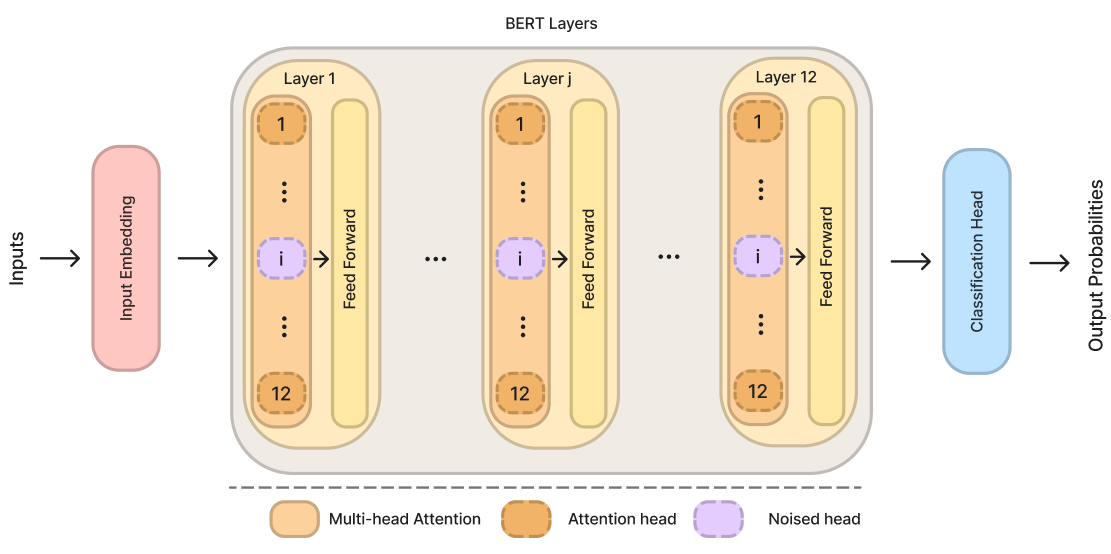}
        \caption{Scheme of head's noising experiments}
        \label{fig:head_noise}
\end{figure} 
 
The encoder then \emph{contextualizes} these representations using a multi-head attention mechanism. The attention mechanism builds attention weights between each pair of words, based on their similarity in a latent space. These are then used to build new word representations by a simple weighted average operation. More precisely, this attention operation is computed for slices of dimensions in parallel before concatenation, this is why it is called multi-head. This is followed by a pass through a feed-forward neural network with residual connections. See Figure \ref{fig:head_noise} for an illustration of this operation.
  The output of the Transformer encoder is a matrix of size $L \times 768$, where $L$ is the maximum document length of the model (512 for BERT) and 768 the number of dimensions. After pre-training on this task, the model can be \emph{fine-tuned} on a downstream task, such as classification, by adding a linear layer on top of this model, that either inputs the final hidden state's "CLS" token 
 (a special token corresponding to a representation of the sequence) or by pooling the representations of the words. 
 
 BERT, in its base version, has 110 M of parameters, 12 layers, and 12 attention heads. DistilBERT is a shallow version of BERT, trained with half the number of layers using distillation \cite{hinton2015distilling}. 
 The principle of this compression method is to train a student model to replicate the behavior of the teacher. 
 To do so, one feeds a dataset to the teacher to retrieve its predictions for each sample (the outputs of the teacher are soft targets, i.e. the probabilities over each class instead of the predicted label). On the other hand, the student receives the same input as training data and the predicted soft targets as training labels. The objective of the student is then to match the (soft) predictions of the teacher. The data used to train the student can either be unlabeled or labeled; in the second case, the true labels can also be fed to the student and a regularization term is added to the objective function to improve the student's performance. Using this approach, DistilBERT obtains up to 95\% of the performance of BERT. 
 
 In our experiments, we use the pre-trained models from Hugging Face\footnote{BERT: \url{https://huggingface.co/docs/Transformers/model\_doc/bert}, \\ DistilBERT: \url{https://huggingface.co/docs/Transformers/model\_doc/distilbert}}, more specifically, the Transformer models for sequence classification with a linear layer as classification head on top of the pooled output.
 
\section{A gender-bias neural exploration protocol for language models}\label{section:protocol}

\subsection{Fine-tuning Scenarios for Fairness Evaluation}
\label{section:models}

In many real-life datasets, we observe gender ratio imbalance, making ML models outcomes prone to unfairness. For example, in the Bias in Bios dataset~\cite{de2019bias}, more than 90\% of nurses are women, while they are less than 2\% to be a surgeon. In \cite{dixon2018measuring}, authors show that training with imbalanced data allows the model to learn a correlation between the target label and sensitive attributes, therefore inducing bias. These observations are unsurprising: this kind of bias is considered to be extrinsic, and is caused by the data used during the fine-tuning. However, with pre-trained models, another type of bias emerges: intrinsic bias. They are encoded during the pre-training and are out of the control of the practitioner. The understanding of the neural behavior that leads to them remains unclear. 
In our work, we are primarily interested in understanding and exploring the inner operations of the Transformer architecture that are at stake in these findings.  

To study in detail the effect of those biases, we fine-tune both BERT and DistilBERT on two sub-samples of the initial dataset: a balanced and an imbalanced one with regard to the sensitive groups (class imbalance remains identical for both datasets). These models will be referred to as \Mone\ and \Mtwo\ when fine-tuned with the imbalanced or the balanced dataset respectively. We believe that starting from the same models, but with different fine-tuning strategies will make it possible to make comparisons of the fairness of these models. 

\subsection{Attention and Hidden States comparison \textbf{(E1)}}
    \label{modelcomparison}
    
    We first investigate the inner differences, induced by the fine-tuning process, between \Mone\ and \Mtwo\ models. 
    (cf. section \ref{section:models}). More precisely, we focus on the attention weights and the hidden states (the intermediate representations) between both models for similar input. 
    For this first set of experiments, we propose to investigate fairness through the lens of the learning dynamic of the PLMs. Recent works \cite{raghu2017svcca} show that the first layers of deep architectures capture low-level information about the input data, and that the learned representations tend to become more abstract and finer when moving through the body of the network towards its heads. In~\cite{hao-etal-2020-investigating}, the authors specifically studied the dynamic of BERT fine-tuning and conclude that mainly the last layers are significantly changing, both their attention mode and the hidden representation that they produce. Our objective is then to verify the two following hypotheses.
    
    \begin{hyp}\label{hyp:1}
    As layers specialize on the different granularity of the textual content, from grammatical to semantic aspects, we assume that monitoring the attention scores and hidden states of the successive layers of the models allows determining which one(s) is encoding bias.
    \end{hyp}
    
    \begin{hyp}\label{hyp:2}

    The distillation process implies that the student model will reproduce the behavior of the teacher, including biases in predictions. We assume that by reducing the depth, and hence the expressive power of the model, compression encourages amplified bias in the hidden representations. 
    \end{hyp}
    
    To proceed, we adapt the protocol of   \cite{hao-etal-2020-investigating} to verify both Hypothesis \ref{hyp:1} and \ref{hyp:2}. We first take a look at the modification of the similarity between tokens, where the attention is computed as a function of the similarity. In a second step, we look at how much hidden word representations are impacted by the model independently of the pairwise similarities.

   \paragraph{\textbf{Attention values comparison.}}
   The Jensen-Shannon divergence is a symmetrized version of the Kullback-Leibler divergence. It allows comparing two probability distributions $P$ and $Q$.
    We propose to compare the attention of the two models layer-wise.  Formally, we evaluate the divergence for each sample and for each head between the layers of two models (e.g., \Mone~ and \Mtwo). Let $N$ be the number of examples in the evaluation set, $H$ the number of attentions head, and $W$ the number of tokens in a sequence. $A^h_i(token_t)$ and $A^h_b(token_t)$ are the attention scores for $token_t$ on head $h$ respectively for models \Mone\ and \Mtwo. In this context, the JS divergence is defined as follows:
     \begin{equation}
        D_{JS}(\Mone||\Mtwo) = \\ \frac{1}{N}\frac{1}{H} \sum_{n=1}^{N}\sum_{h=1}^{H} \frac{1}{W} \sum_{t=1}^{W} D_{JS}(A^h_i(token_t)||A^h_b(token_t))
    \end{equation}
    $D_{JS}(.||.) \in [0, 1]$, where $0$ indicates that the distributions are identical.  
    
    \paragraph{\textbf{Hidden states comparison.}}
     We compute the Singular Vector Canonical Correlation Analysis distance (SVCCA)~\cite{raghu2017svcca} to observe the evolution of hidden states. 
     SVCCA allows analyzing and comparing representations in deep learning models; in our case, the hidden representations produced by each Transformer layer. When computing SVCCA, we first perform a Singular Value Decomposition (SVD) of the representations produced by the two models for each input observation. Then, we compute the Canonical Correlation Analysis (CCA) \cite{hardoon2004canonical} between the two subspaces created by the SVD to evaluate the correlation between the two representations and finally, condense the correlations obtained for each dimension into a distance.  

     Let $c$ be the hidden size of the model and $\rho \in [0, 1]$ the CCA. The SVCCA distance is defined as follows:
    \begin{equation}
        D_{SVCCA}(\Mone||\Mtwo) = 1 - \frac{1}{c} \sum_{j=1}^{c} \rho^{(j)}
    \end{equation}
    
     $D_{SVCCA}(.||.) \in [0, 1]$ with  0 meaning identical representations.
     
\subsection{Head's ablation \textbf{(E2)}}
\label{headnoising}
    With multi-head attention, Transformers build for each head a different representation of the input embedding. We make the hypothesis that some representations might induce more biases than others. 
    Thus, complementary to the previous experiments and in continuity with the goal of finding where are biases encoded in PLM, we successively ablate heads of the model to infer if some of them are responsible for biases in the model. The ablation is done by setting all its attention weights to 0 through all the layers, as shown in Figure \ref{fig:head_noise}.
    \begin{hyp}\label{hyp:3}
    By ablating attention heads, we aim at removing the bias due to a given head and identify the ones contributing to unfairness. In other words, we expect that when ablating a head responsible for bias, the model will obtain a better fairness score and reciprocally. 
    \end{hyp}

    In practice, we first fine-tune the model so that it learns the weights as in real-world applications. Then, we successively ablate heads and evaluate the performance and fairness of the model on new data to evaluate the bias encoded by the aforementioned heads.
    We are interested in the results of our models following two criteria: their predicting performance and their fairness. 
    
    \paragraph{\textbf{Performance.}} The model performance is evaluated using a weighted version of the F-Score (since our target variable is multivalued). 
    
    More precisely, we compute the F-Score for each class, then compute the average weighted by the number of samples per class. 

    \paragraph{\textbf{Gender fairness.}} We are interested in group fairness, and several metrics have been proposed in the literature \cite{caton2020fairness}. We choose the commonly used Equalized Odds (EO) \cite{hardt2016equality}, defined as follows $\mathbb{P}(\overline{y}=1 | y, S = 0) = \mathbb{P}(\overline{y}=1 | y, S = 1)$. \\
    To ease the interpretation, we compute the difference version of EO given by 
    \begin{equation}
        EO = |\mathbb{P}(\overline{y}=1 | y, S = 0) - \mathbb{P}(\overline{y}=1 | y, S = 1)|,
    \end{equation}
    where $\overline{y}$ are the predictions, $y \in {0, 1}$ are the true labels and $S$ corresponds to the sensitive attributes (0 and 1 indicating the belonging to a sensitive group). EO $\in$ [0, 1] where a score closer to 0 indicates fairer predictions.

\section{Experiments}
\label{section:experiments}

\subsection{Task and Dataset}
    \label{section:dataset}
     In our experiments, we focus on a classification task and use a subset of the Bias in Bios dataset \cite{de2019bias} called the Curriculum Vitae dataset\footnote{Dataset: \url{https://www.kaggle.com/competitions/defi-ia-insa-toulouse/data}}. It contains a set of short biographies associated with an occupation and a gender. The dataset is composed of 217,197 entries, and $28$ professional occupations. The distribution of classes (occupations) and groups (genders) within each occupation is highly imbalanced.
     For example, class 19 corresponding to `professor' represents 32.23\% of the dataset and within the class, women represent 44.88\% of the entries; class 23 corresponding to `paralegal' represents 0.44\% of the dataset, and women 84.17\% of the class entries.
     
     We create two versions of this dataset, a balanced version, and an imbalanced version, depending on the gender attribute. The former one is a subset, where for each class the largest sensitive group is truncated to equalize the proportion of individuals of each gender. The latter is a subset, where we reproduce the imbalance between gender observed in the initial dataset, but both groups are truncated to ensure that the number of samples in both subsets are equal. 
     
     Based on these two versions, we exploit the relationship between fairness and gender imbalance (cf. section \ref{section:models}) to build two models \Mone\ and \Mtwo\ to further explore the mechanisms of bias. We evaluate the EO of BERT and DistilBERT fine-tuned on $70\%$ of the samples, for both versions of the original dataset.
     In the sequel, we refer to these models as \B\_\Mtwo~and \DB\_\Mtwo~for the balanced versions, and \B\_\Mone~and \DB\_\Mone\ for the imbalanced ones. To confirm our premise, we perform classification and observe an average EO over all classes three times higher for the imbalanced versions (0.13 vs. 0.42). 
     Following this first experience, one might think that balancing the fine-tuning data is a sufficient and satisfactory solution to ensure fairness. However, before proceeding further, two remarks are in order. First, balancing the data is a first step in reducing bias, but it does not guarantee a fair model (EO above $0$). Second, in many real-world scenarios, where multiple protected attributes can be observed simultaneously (e.g. women of color), this solution appears to be shortsighted as one cannot slice the data into more sub-population infinitely to rebalance classes. 

    \begin{figure}[t]
      \centering
      \includegraphics[width=.24\textwidth]{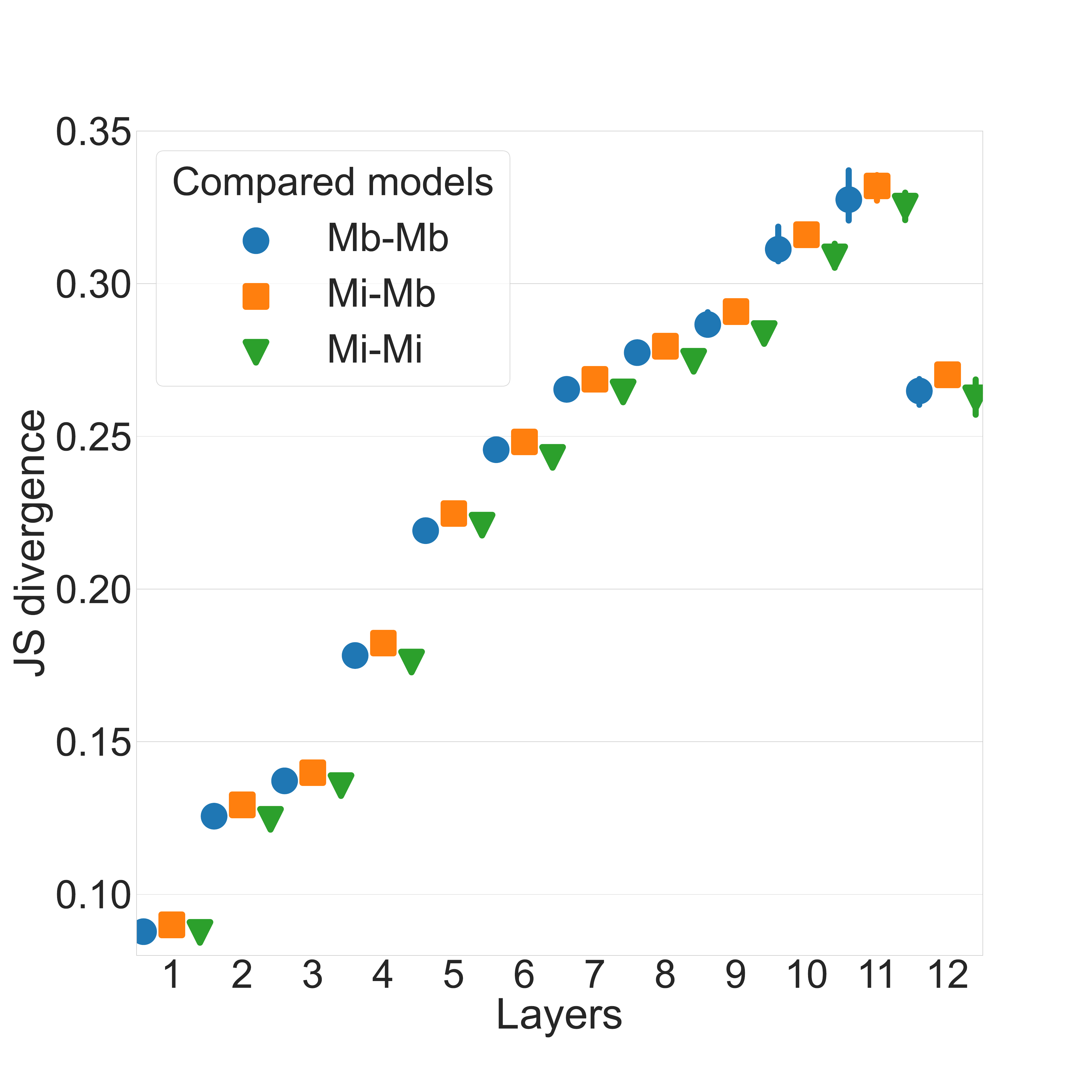}
      \includegraphics[width=.24\textwidth]{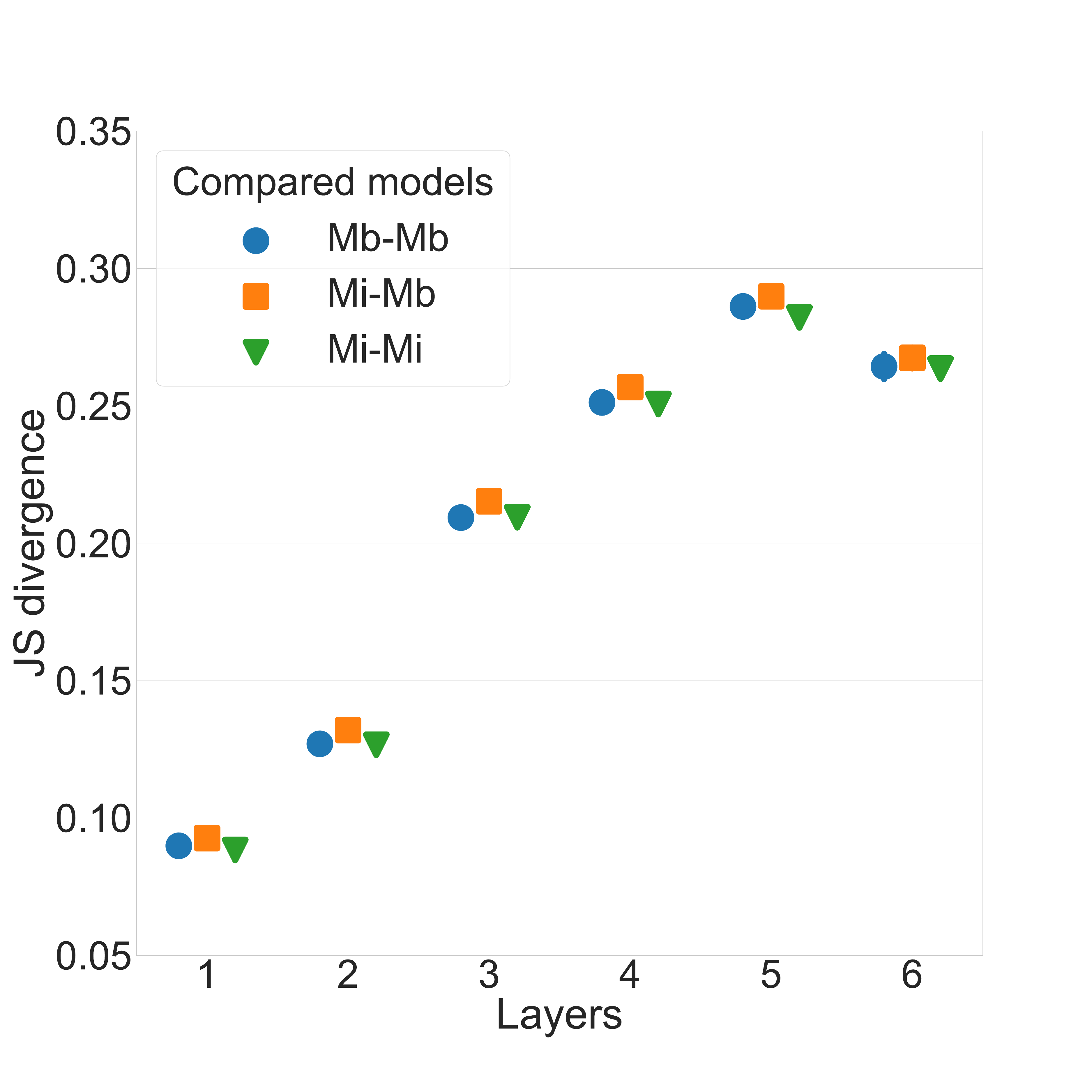}
      \includegraphics[width=.24\textwidth]{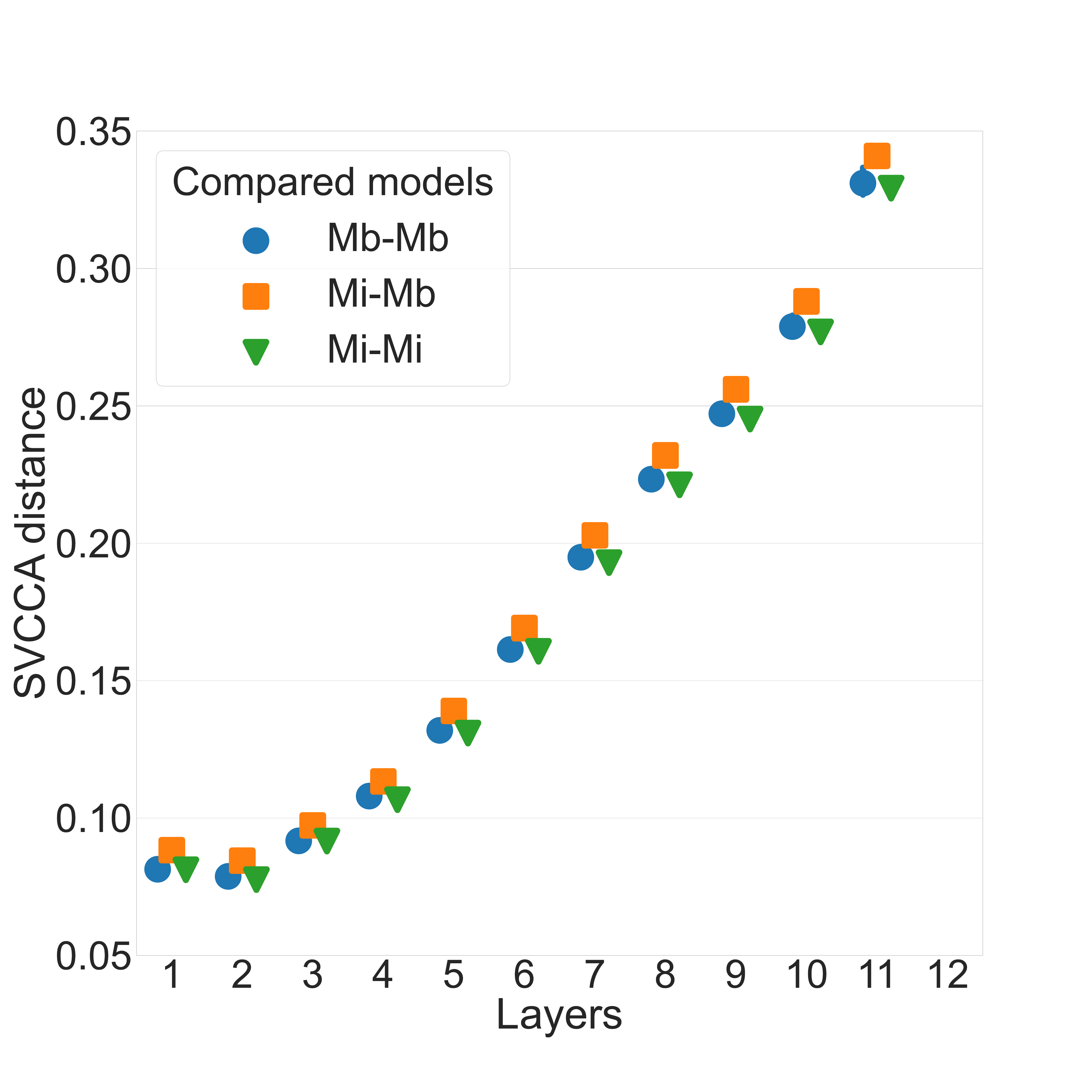}
      \includegraphics[width=.24\textwidth]{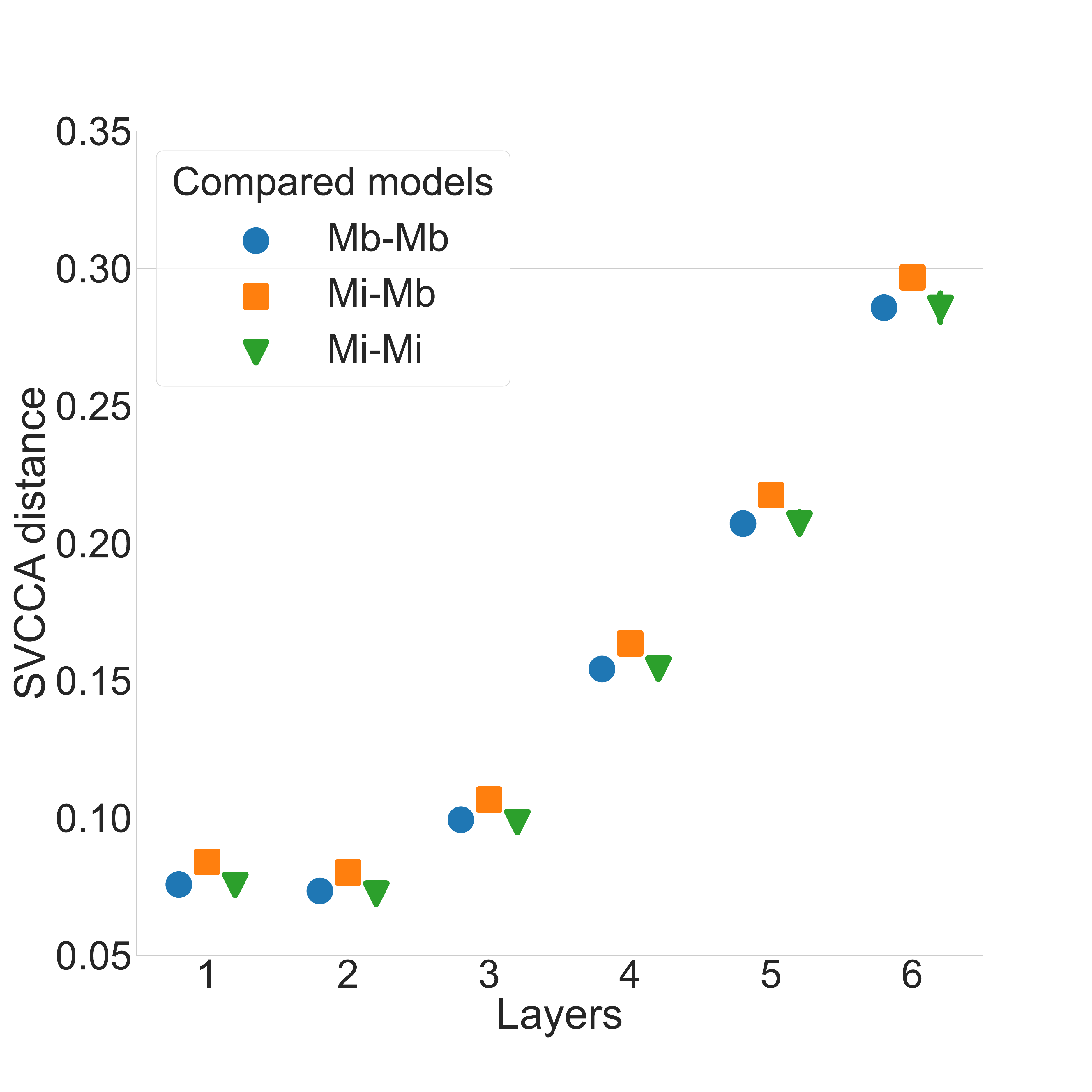}
        \caption{JS divergence comparison for BERT (left) and DistilBERT (center-left) and SVCCA distance for BERT (center-right) and DistilBERT (right).}
        \label{fig:js_svcca}
    \end{figure}
        
\subsection{Attention and Hidden States comparison \textbf{(E1)}}
	We present the results of this experiment averaged over five random seeds in Figure \ref{fig:js_svcca}. 

	If we observe higher divergence between models \Mone \ and \Mtwo \ than between \Mone \ and \Mone \ or \Mtwo \ and \Mtwo \ on a given layer, we can assume this layer to be responsible for encoding bias. 
	For the JS divergence, first we can note a similar pattern for both BERT and DistilBERT: the divergence increases as we move forward in the architecture, with a peak on the penultimate layer. For the 
    SVCCA distance, the trend is similar, and we reach the highest value on the last layers. These findings are perfectly in line with the results of \cite{hao-etal-2020-investigating} and \cite{merchant2020happens} claiming that fine-tuning mainly affects top layers. 
    These first observations allow us to state that DistilBERT follows the same learning dynamic than BERT during fine-tuning. 
    Now, comparing [\Mone-\Mone, \Mtwo-\Mtwo] vs. \Mone-\Mtwo, we observe that the values for both metrics are slightly above on \Mone-\Mtwo, but not significantly. In addition, this difference is consistent over all layers. Finally, we observe the exact same behavior for DistilBERT.
    These particular results are counterintuitive with our Hypotheses \ref{hyp:1} and \ref{hyp:2}, and we cannot conclude that extrinsic bias makes some layers different with regard to internal representations and attention scores for both architectures.
    
    \begin{figure}[t]
        \centering
        \subfloat[]{\includegraphics[width=.49\textwidth]{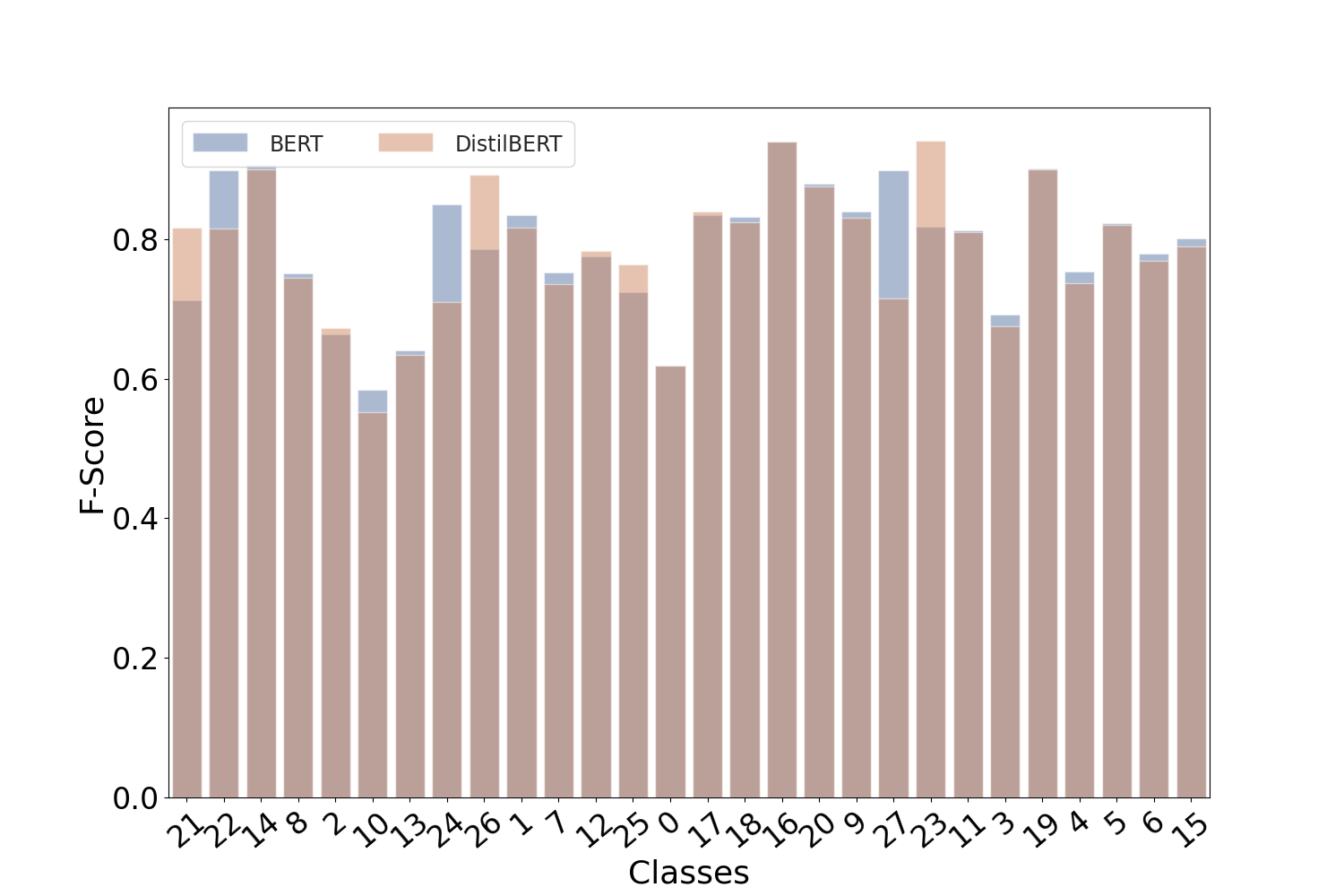}}
        \subfloat[]{\includegraphics[width=.49\textwidth]{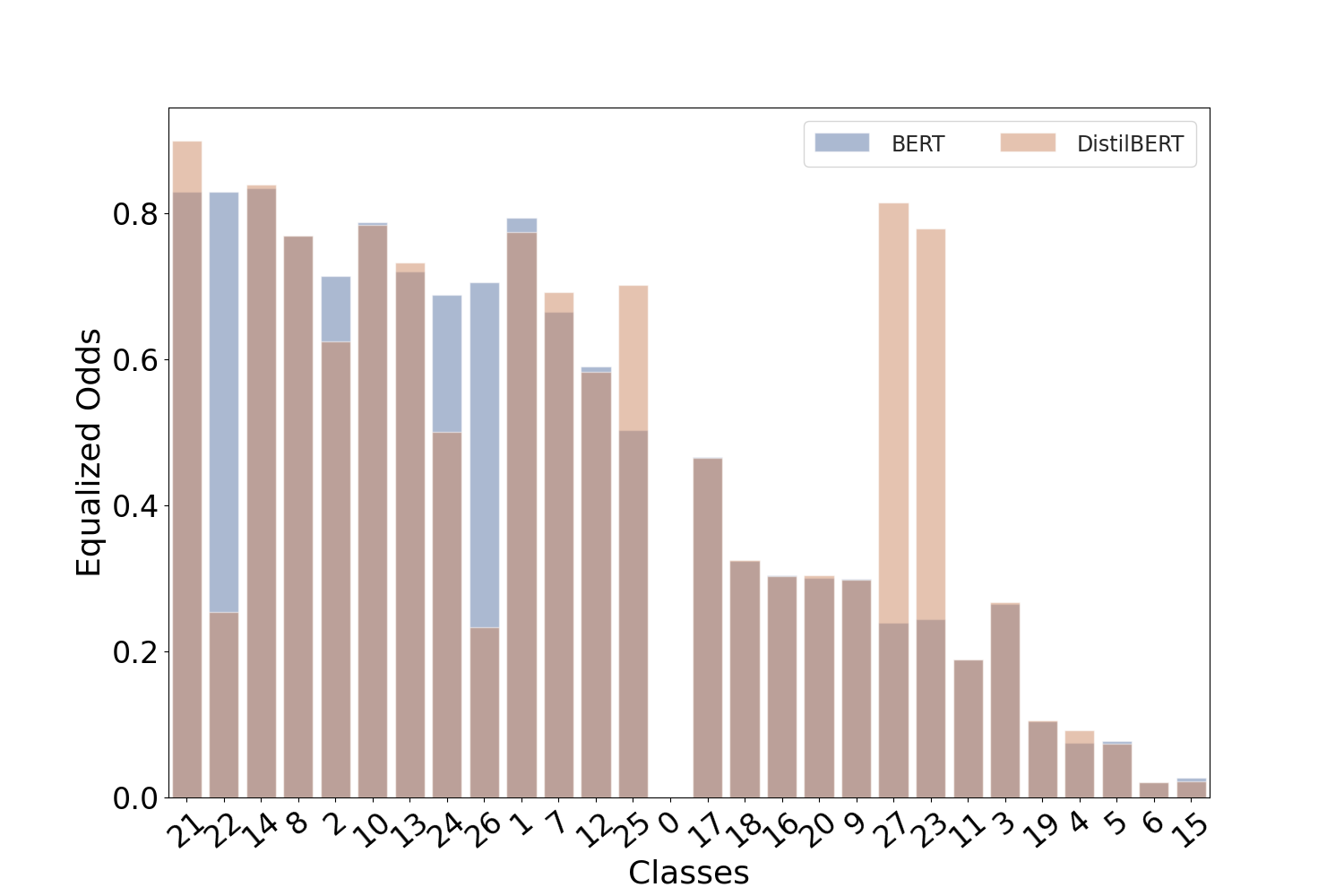}}
        \caption{(a) F-Score and (b) Equalized Odds for model \Mone\, per class, without ablation and ordered by ascending ratio W/M.}
        \label{fig:score_without_noise}
    \end{figure}
    
\subsection{Head's ablation \textbf{(E2)}}

\label{sec:heads}
    Let us now look at the relationship between the heads and bias. Once the model is fine-tuned, we neutralize the heads in turn at inference time to estimate the bias they encode. Since the distribution of the classes is highly imbalanced, we evaluate the fairness of the model for each class individually. \\
    Figure \ref{fig:score_without_noise} shows the EO and F-Score for each class on the original model (\Mone) (classes are sorted by ascending ratio women/men). Three observations can be drawn: firstly, for the highly imbalanced class (left-learning) EO is significantly higher than average; secondly, comparing BERT and DistilBERT, we see that for the majority of classes, we obtain an equivalent level of fairness, except for a few classes, where either one or the other is more biased. However, for the most imbalanced classes, DistilBERT is reaching a better level of fairness in comparison with BERT (lower or equal EO scores); thirdly, both architectures obtain comparable F-Score. 
    
\begin{figure}[t]
        \centering
        \subfloat[]{\includegraphics[width=.32\textwidth]{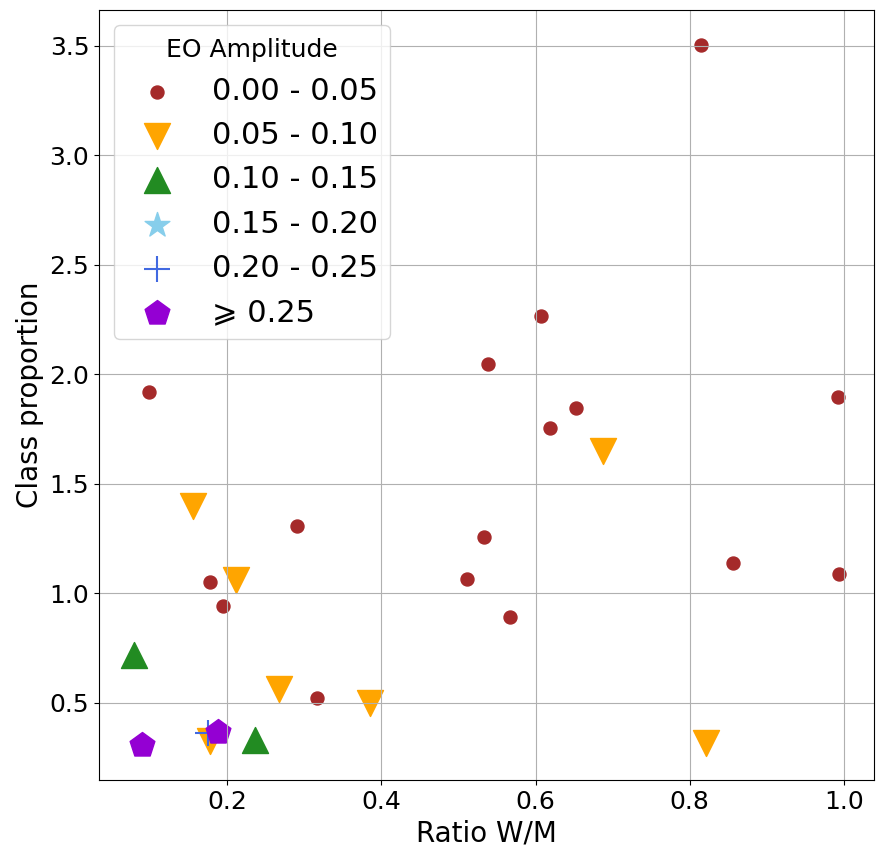}} \label{fig:eo_amplitude1}
        \subfloat[]{\includegraphics[width=.32\textwidth]{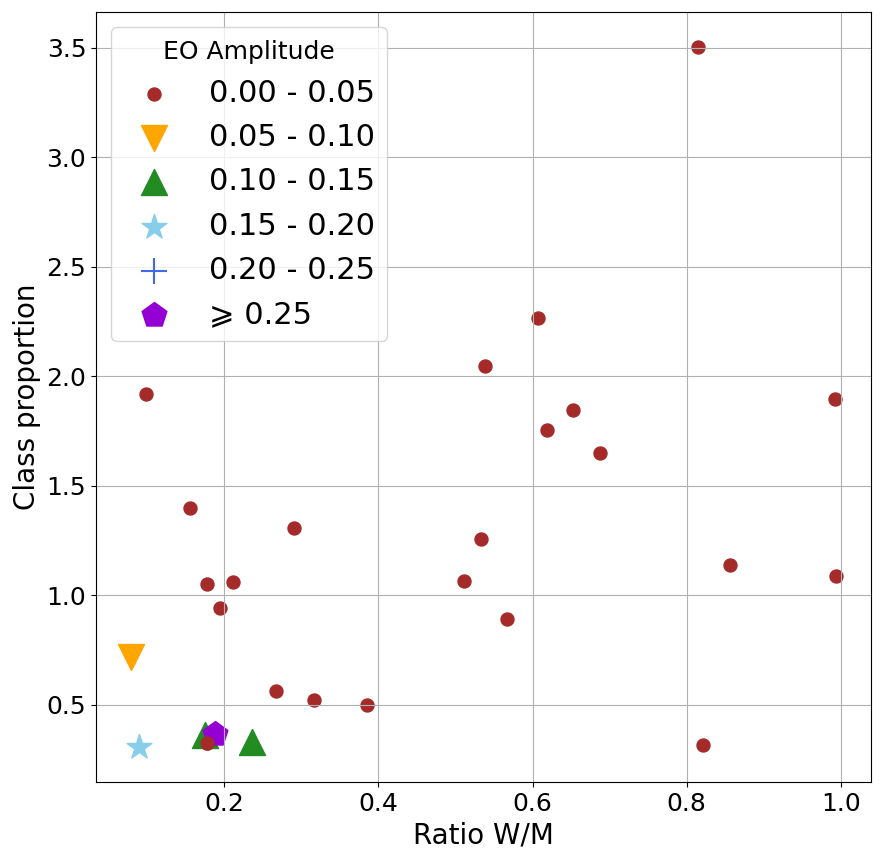}} \label{fig:eo_amplitude2}
       \subfloat[]{\includegraphics[width=.32\textwidth]{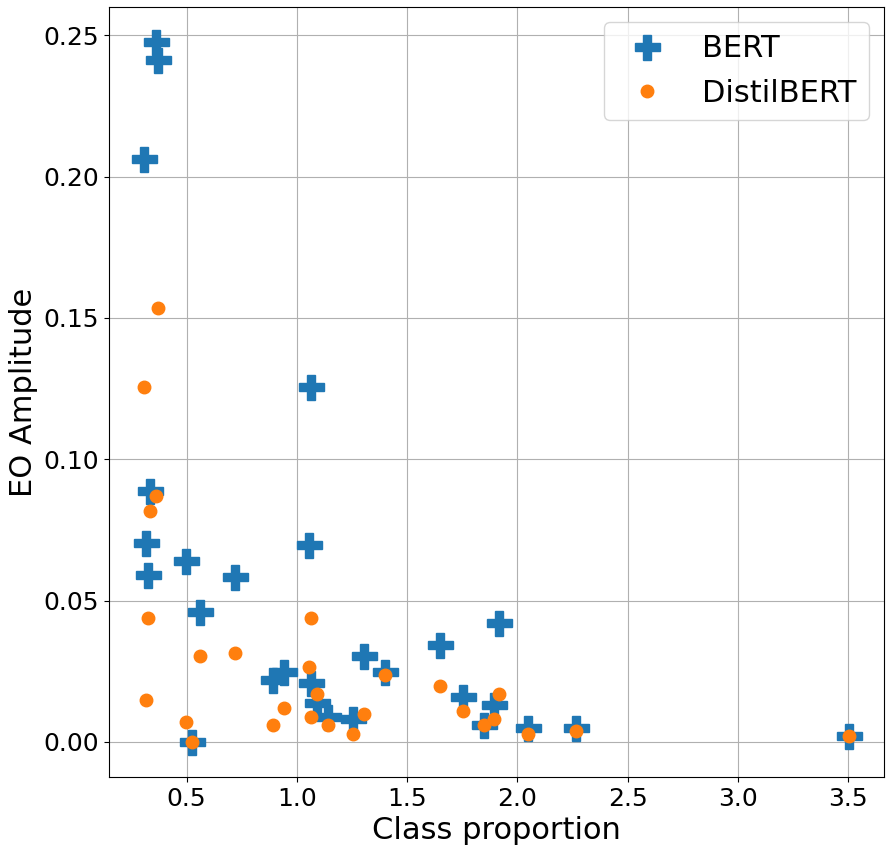}} \label{fig:eo_amplitude3}
        \caption{EO amplitude for BERT's \Mone\ (a), DistilBERT's \Mone\ (b), BERT's and DistilBERT's
        \Mtwo\ (c).}
        \label{fig:eo_amplitude}
    \end{figure}

    Now, we reproduce this evaluation twelve times (one time for each ablation), and observe different levels of variations for the EO depending on both the class and the head that is ablated. This implies that the representations produced by the attention head are different enough to impact the fairness of the models as assumed in Hypothesis 3. However, when the network is re-fine tuned we do not observe the same variations for a given head. For each class, we compute the difference of EO obtained when neutralizing a head gives the fairest score vs. the most unfair score and call it the amplitude. Let $EO_{class} = \{EO^{1}_{class},...,EO^{12}_{class}\}$ with $EO^{head}_{class}$ the EO computed for a given class after noising a head. We defined the amplitude for a class as :
    \begin{equation}
        \text{amplitude}_{class} = \max_{head}(EO_{class}) - \min_{head}(EO_{class})
    \end{equation}
    Figure \ref{fig:eo_amplitude} shows the amplitude depending on the class imbalance and the ratio of women/men within the classes. The minority group is not the same for every class, thus, we compute the ratio as follows $\text{ratio W/M} = \min(\frac{\%women}{\%men}, \frac{\%men}{\%women}) \ \in [0, 1]$.
    For better readability, we rescale the class proportion and amplitude vectors by taking $\text{log}(\text{vector} + 1_n)$, $n$ being the dimension of the vectors.
    We note, in Figures \ref{fig:eo_amplitude1} and \ref{fig:eo_amplitude2} that the more a class suffers from double imbalance (underrepresented class and highly sensitive group imbalance) the more the heads will produce different representations, some being more biased than others. Also, when comparing the values of EO amplitude for BERT and DistilBERT, we observe that BERT is more sensitive to those scenarios than DistilBERT. According to Figure \ref{fig:eo_amplitude3}, where sensitive groups are balanced within the classes, the less a class is represented the higher the amplitude is, meaning that \textbf{BERT is generally more sensitive to class imbalance than DistilBERT with regard to the homogeneity of head representations}. On the other hand, we evaluate the correlation between F-Score and EO, when ablating each head, and have not been able to establish a relation caused by the process.
    
\section{Conclusion}
\label{section:conclusion}
    This paper investigated the implication of inner elements of BERT-based models' architecture in bias encoding through empirical experiments on the Transformers' layers and attention heads. 
    We also studied the attention carried by the "CLS" token to the words of the sequence, specifically the pronouns 'he' and 'she', but also to the ones receiving the most attention from the aforesaid token, in an attempt to understand what the model was focusing on; this study did not lead us to any convincing results. Similarly, investigating the JS divergence and SVCCA distance between different layers (e.g. the 1 and 2) was not conclusive, we suspect it might since layers specialize on different aspects of the input text \cite{jawahar2019does}.
    To summarize, we show that gender bias is not encoded in a specific layer or head. We also demonstrate that the distilled version of BERT, DistilBERT, is more robust to double imbalance of classes and sensitive groups than the original model. Even more specifically, we observe that the representations generated by the attention heads in such a context are more homogeneous for DistilBERT than for BERT in which some attention heads will be fair while others are very unfair.
    Thus, we advise giving special care to such patterns in the data but do not recommend ablating the heads producing more unfair representations since it could seriously harm the performance of the model. Finally, we recommend DistilBERT to the practitioner using datasets containing underrepresented classes with a high imbalance between sensitive groups, while cautiously evaluating class independently, using the protocol that we propose in this paper. 
    
\section*{Acknowledgment} This work is funded by the french National Agency for Research (ANR) in the context of the Diké project (ANR-21-CE23-0026).

%
%
%
\bibliographystyle{splncs04}
\bibliography{biblio}

\end{document}